\documentclass[11pt,a4paper,twocolumn,journal,a4paper]{IEEEtran}
\usepackage{cite}
\usepackage{amsmath,amssymb,amsfonts}
\usepackage{algorithmic}
\usepackage{graphicx}
\usepackage{textcomp}
\usepackage{caption2}

\onecolumn

\def\BibTeX{{\rm B\kern-.05em{\sc i\kern-.025em b}\kern-.08em
    T\kern-.1667em\lower.7ex\hbox{E}\kern-.125emX}}
\begin{document}
%\history{Date of publication xxxx 00, 0000, date of current version xxxx 00, 0000.}
%\doi{10.1109/ACCESS.2017.DOI}

\title{An Unified Intelligence-Communication Model for Multi-Agent System Part-I: Overview}
\author{Bo Zhang, Bin Chen, Jinyu Yang, Wenjing Yang, Jiankang Zhang
\IEEEcompsocitemizethanks{\IEEEcompsocthanksitem
B. Zhang is with the Artificial Intelligence Research Center, National Institute of Defense Technology Innovation, Beijing, P. R. China. E-mail: bo.zhang.airc@outlook.com.

B. Chen is with the Academy of Military Sciences, Beijing, P. R. China. J. Yang and W. Yang are with the State Key Laboratory of High Performance Computing, National University of Defense Technology, Changsha, P. R. China. J. Zhang is with the Southampton Wireless Group, Southampton University, SO17 1BJ, the United Kingdom.
%\IEEEcompsocthanksitem Corresponding: Dr. Wenjing Yang. E-mail: wenjing.yang@nudt.edu.cn
}}
\maketitle
\vspace{-2cm}

\begin{abstract}
Motivated by Shannon's model and recent rehabilitation of self-supervised artificial intelligence having a "World Model", this paper propose an unified intelligence-communication (UIC) model for describing a single agent and any multi-agent system.

Firstly, the environment is modelled as the generic communication channel between agents. Secondly, the UIC model adopts a learning-agent model for unifying several well-adopted agent architecture, e.g. rule-based agent model in complex adaptive systems, layered model for describing human-level intelligence, world-model based agent model. The model may also provide an unified approach to investigate a multi-agent system (MAS) having multiple action-perception modalities, e.g. explicitly information transfer and implicit information transfer.

This treatise would be divided into three parts, and this first part provides an overview of the UIC model without introducing cumbersome mathematical analysis and optimizations. In the second part of this treatise, case studies with quantitative analysis driven by the UIC model would be provided, exemplifying the adoption of the UIC model in multi-agent system. Specifically, two representative cases would be studied, namely the analysis of a natural multi-agent system, as well as the co-design of communication, perception and action in an artificial multi-agent system. In the third part of this treatise, the paper provides further insights and future research directions motivated by the UIC model, such as unification of single intelligence and collective intelligence, a possible explanation of intelligence emergence and a dual model for agent-environment intelligence hypothesis.

\textit{Notes: This paper is a Previewed Version, the extended full-version would be released after being accepted.}
\end{abstract}

\begin{IEEEkeywords}
Information Theory, Multi-Agent System, Learning (artificial intelligence), Communication Channels.
\end{IEEEkeywords}

\section{A BRIEF INTRODUCTION OF THE UIC MODEL - FROM SHANNON'S PERSPECTIVE}
\label{sec:introduction}
This Section introduces the motivation of the UIC model and the basic graphical description of the UIC model, illustrating the relationship between an intelligent agent and the environment. This section would cover the following aspects.
\begin{itemize}
\item The environment is model as the generic communication channel between agents.
\item The agent is model as an communication transceiver. Specifically, the internal world model is model as a source coder, which encodes the environment, while the multi-modal perceiving and acting capability are abstracted as multiple transmitter-receiver pairs.
\end{itemize}

Before exposing the proposed UIC model, let us first review the Shannon Model.

\section{A BRIEF REVIEW OF THE SHANNON MODEL}
This section applies the Shannon Model to describe complex forms of interactions between agents and environments.

The Shannon model proposed in the pioneering paper “A mathematical theory of communication” may be illustrated as follows, and please note that the Weiner’s feedback is not included for now.

\begin{figure}[h]
\centering
\includegraphics[width=0.9\linewidth]{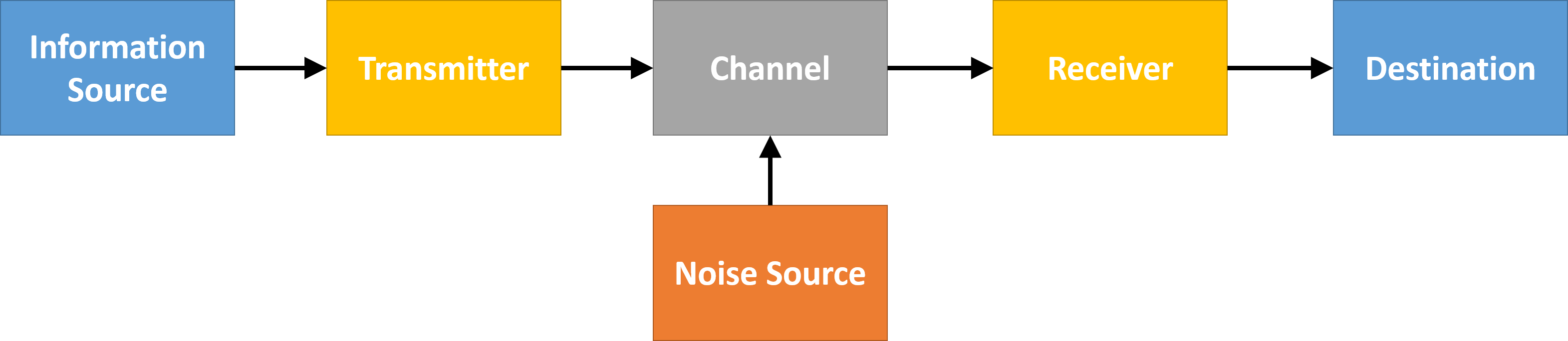}
\caption{Shannon Model without Feedback [1]}
\label{fig1}
\end{figure}

\section{COMPLEX INTERACTIONS MODELING: IS SHANNON MODEL SUFFICIENT?}
This paper is definitely NOT the first effort to apply Shannon Model to model and understand the interactions between agents and the environment.

An intuitive question arose immediately (especially the critics from the communication society and most of the control society) and therefore may prohibit the further efforts. Is the Shannon channel model sufficient to describe complex interactions between agents and the environment? As an agent not only transmits and receives waveforms, but also changes the physical environment, e.g. moving objects, absorbs solar energy, etc.

Before answering the above question directly, the author may redirect to a more fundamental question: \textbf{What is the relationship between Information, Substance and Energy?} The answer to this question may help determine the range of applications of Shannon Channel Model. Hence, the UIC model may be firmly built based on the following statements:
\begin{itemize}
\item Information is carried by waves. The waves may be electromagnetic waves, mechanical waves, matter waves and other forms of waves, which are generated by the four fundamental interactions.
\item The environment may be modeled as a possibly infinite set of waves. Each wave carries a certain amount of information and energy. The different forms of waves may interact, exchanging energy and information.
\item Any form of interactions may be contaminated by noise or uncertainty. Therefore, the maximum amount of information that may be transferred by a wave is quantified by the Shannon Channel Capacity.
\end{itemize}

Based on the above statements, we may come to the following statements: The environment may be model as a set of multi-modal channels for different forms of waves.

\section{THE AGENT MODEL IN THE UIC MODEL}
After modelling the environments, and based on the widely adopted philosophy that an intelligent agent is an open system in the environment, the UIC model incorporates the most important assumptions of an intelligent agent.

The UIC Assumption: As with the Environment, an agent is also a set of waves, and the interactions between an agent and the environment may be model as interactions between waves. The interaction process exchanges energy and information, meanwhile incorporates uncertainties.

Therefore, the interactions between an agent and the environment may be categorized by the widely used terminologies in describing the behaviour of an intelligent agent:

\subsection{Agent Perception}
Perception may be considered as the measurement of energy and information carried by a wave in the environment.
\begin{itemize}
\item Based on the UIC Assumption, perception is modeled as the energy and information transferring from the environment to the agent.
\item The perception module of an agent is then model as a set of channel receivers, where each receiver extracts information from the waves, and an optimal receiver extracts the maximum amount of information, as bounded by the Shannon Capacity.
\end{itemize}

\subsection{Agent Action}
Action taken by an agent may be in the macro and micro level, and in various domains, e.g. spatial, frequency, energy, etc.
\begin{itemize}
\item Based on the UIC Assumption, any action may be model as a form of wave interactions, action is model as the energy and information transferring from the agent to the environment.
\item The action module of an agent is then model as a set of channel transmitters, where each transmitter modulates a certain set of waves with information, and a feedback from the receiver, as shown by Shannon and Wiener, may greatly improve the optimal transmitter-receiver design.
\end{itemize}

We may now graphically depict the first UIC model figure, focusing on using the Shannon model to interpret the interactions between the agent and the environment.

\begin{figure}[ht]
\centering
\includegraphics[width=0.3\linewidth]{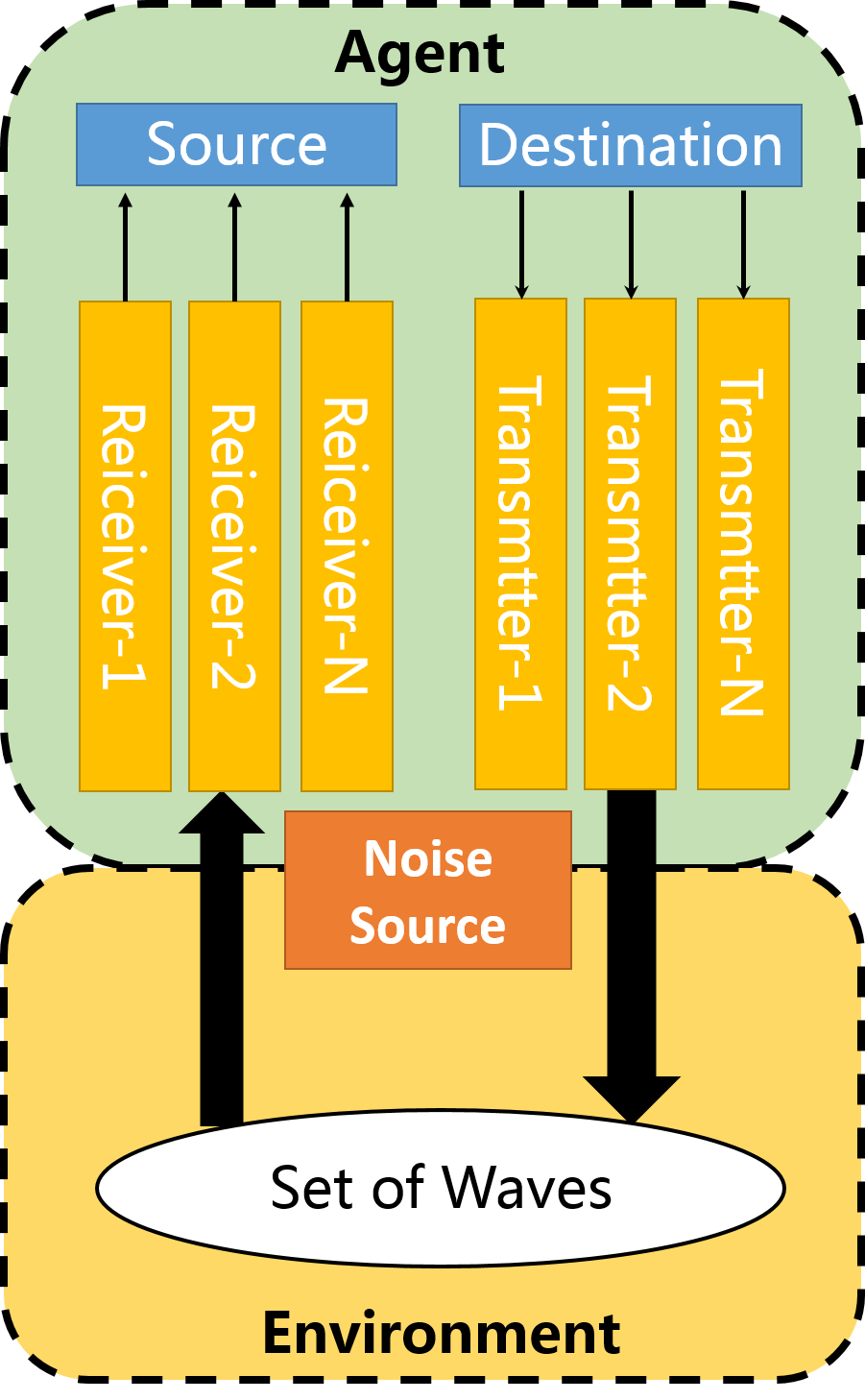}
\caption{Interactive Model between the Agent and the Environment}
\label{fig2}
\end{figure}

In the up-coming sections, we will dive deeper into the information source and destination in the agent.

\section{MODELLING THE INFORMATION SOURCE AND DESTINATION IN THE AGENT}
This Section introduces the model for information source and destination in the UIC model as proposed in the last section. This section would cover the following aspects.

\begin{itemize}
\item The information source and destination is modelled, following the generic learning agent model, which includes a performance element, a learning element, a critic and a problem generator.
\item This Section reviews the different agent models, with an emphasis on the design of learning agent model, and propose an unified, complete and operable design for the learning agent that may lead to strong AI.
\item Along with the perception and action model proposed in the last section, we may form the complete UIC model for an intelligent agent.
\end{itemize}

If we stick to the Shannon's model of communication, the information source and destination may be disconnected in a single agent. However, in order to allow an agent to receive information from the outside world, to process information and to change the world by transmitting information, a connection between the information source and destination in the agent is inevitable. Therefore, a question arose: How does an intelligent agent process the information within itself? This paper is definitely NOT the first to answer the question, so let us review the agent internal models.

\section{A BRIEF REVIEW OF THE AGENT INTERNAL MODELS}
An overview of the major five categories of agent models is given in Section 2.4 of [2]. Except for the simplest form of agent model, the rest of the agent models includes an internal world model.

\subsection{Agent Models without Learning}
The agent without learning may be modelled as follows.
\begin{itemize}
\item Simple Reflex Agent model (SRA model): The agent function is based on the condition-action rule: if condition then action. So it may work well in a fully-observable world.
\item Model-based, Reflex Agent model (MRA model): A world model is introduced to maintain some kind of structure, which describes the part of the world which cannot be observed.
\item Model-based, Goal-Based Agent model (MGA model): The goal information is introduced to describe desirable situations and allow an agent to choose among multiple possibilities, selecting the one which reaches a goal state. In the complex adaptive system, Holland proposed to incorporates co-existence of multiple micro SRA agents in an agent, and allowing the agent to select action sequences[3].
\item Model-based, Utility-Based Agent model (MUA model): The utility function is introduced to measure the space spanning from the goal state and the non-goal state.
\end{itemize}

\subsection{Learning Agent Models}
Afore-mentioned agent models are static, as they do not explain how an agent may learn and evolve. Therefore, the LA model is quite different by introducing four conceptual components:
\begin{itemize}
\item A performance element to select external actions, which is equivalent to a static agent model that percepts from the world and decides external actions.
\item A learning element to make improvements by changing the performance element and gathering knowledge from the performance element.
\item A critic to enable feedback on how the agent is doing and how the performance element may be modified to do better.
\item A problem generator to select actions leading to new and informative experiences.
\end{itemize}

\begin{figure}[ht]
\includegraphics[width=0.7\linewidth]{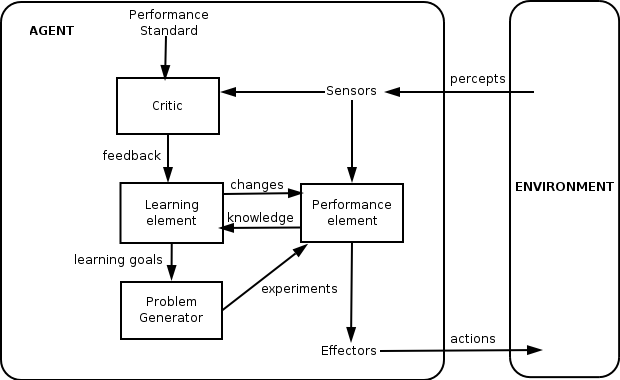}
\centering
\caption{Learning Agent model (LA model) [2]}
\label{fig3}
\end{figure}

\section{RETHINK THE LEARNING AGENT MODEL WITH AN INTERNAL WORLD MODEL}
Although it is not pointed out explicitly in [2], the learning agent model may include a world model in its performance element. Therefore, the learning element may change the world model.

If the design principle of the learning agent model is accepted (why NOT if we have not come up with a more reasonable model), then the problem of designing an intelligent agent may be further decomposed as follows:

\begin{itemize}
\item Performance Element Design: What is inside the performance element and what is a WM looks like?
\item Learning Element and Critic Design: How to gather knowledge and improve performance based on the feedback provided by the critic in a complex dynamic world? Also, how to update the WM and rest of Performance Element?
\item Problem Generator Design: How to find a GOOD tradeoff between short-term and long-term reward in a changing world?
\end{itemize}

The answers to the above four questions may lead to the design principles of Strong Artificial Intelligence. Therefore, the answers to these questions are so attractive and many pioneering thoughts (NOT solutions) have been proposed to light our way.

\subsection{Agent Model in Complex Adaptive Systems}
Professor Holland H. John is the pioneer of Complex Adaptive System Theory. In his 1995 treatize [3], Holland proposed the following works:
\begin{itemize}
\item Performance Element Design: Holland proposed the design of the performance system, which has an input from sensors, an output to the actuators, while the performance element is a collection of interactive micro-agents based on IF-THEN rules.
\item Learning Element and Critic Design: A credit assignment mechanism was proposed to allow multiple micro-agents to compete with each other. The micro-agent with higher reward from the critic is given a larger credit, hence improving performance of the agent.
\item Problem Generator Design: A rule discovery mechanism was proposed based on the genetic algorithm, which was also invented by Holland.
\end{itemize}

\subsection{Agent Model with Commonsense Thinking}
In [5], Professor Marvin Minsky proposed to model a (human) brain into a collection of resources, which may be activated by different emotion states. The thinking activities may be broken down into a series of six-layered actions: instinctive reactions, learned reactions, deliberate thinking, reflective thinking, self-reflective thinking and self-conscious reflection. Higher-level thinking is built upon the ones below, and the lowest instinctive reaction level is driven by If-Do rules.

Specifically, we may re-organize Minsky's thoughts as follows:

\begin{itemize}
\item Performance Element Design: The instinctive reactions may be modelled as IF-DO rules, which are equivalent to Holland's IF-THEN rules.
\item Learning Element and Critic Design: The higher-layer actions are built upon instinctive reactions. Minsky emphasized the limitation of IF-THEN rules because the scalability of directly using IF-THEN rules for describing the complex world is infeasible, therefore, representations, evaluations and manipulations of higher-level abstraction is required. It should be noted that Holland also dived much deeper later on following the same rule of agent model, introducing the concept of dynamic finitely generated systems for modelling complex interactions between agents [3].
\item Problem Generator Design: The layers higher than deliberate thinking motivate an agent to criticize the reactive actions and explore un-explored world.
\end{itemize}

\subsection{Self-Supervised Learning Agent Model}
Very recently, Professor Yann Lecun proposed an agent model based on the concept of Self-Supervised Learning with both a deep structure and a world Model[5]. The agent may do reasoning by combining predicting and planning with the aid of model-based deep reinforcement learning. Specifically, Lecun considered the following strategies:
\begin{itemize}
\item Performance Element Design: A world model is needed, accumulating background knowledge about how the world works and perhaps include common sense. Deep (artificial neural) networks may model complex interactions with the aid of non-linear transformations within a manageable scale.
\item Learning Element and Critic Design: Latent-variable forward models is proposed for planning and learning policies, but it is also far from a complete solution.
\item Problem Generator Design: It seems Lecun did not consider it for the moment.
\end{itemize}

\begin{figure}[ht]
\includegraphics[width=0.7\linewidth]{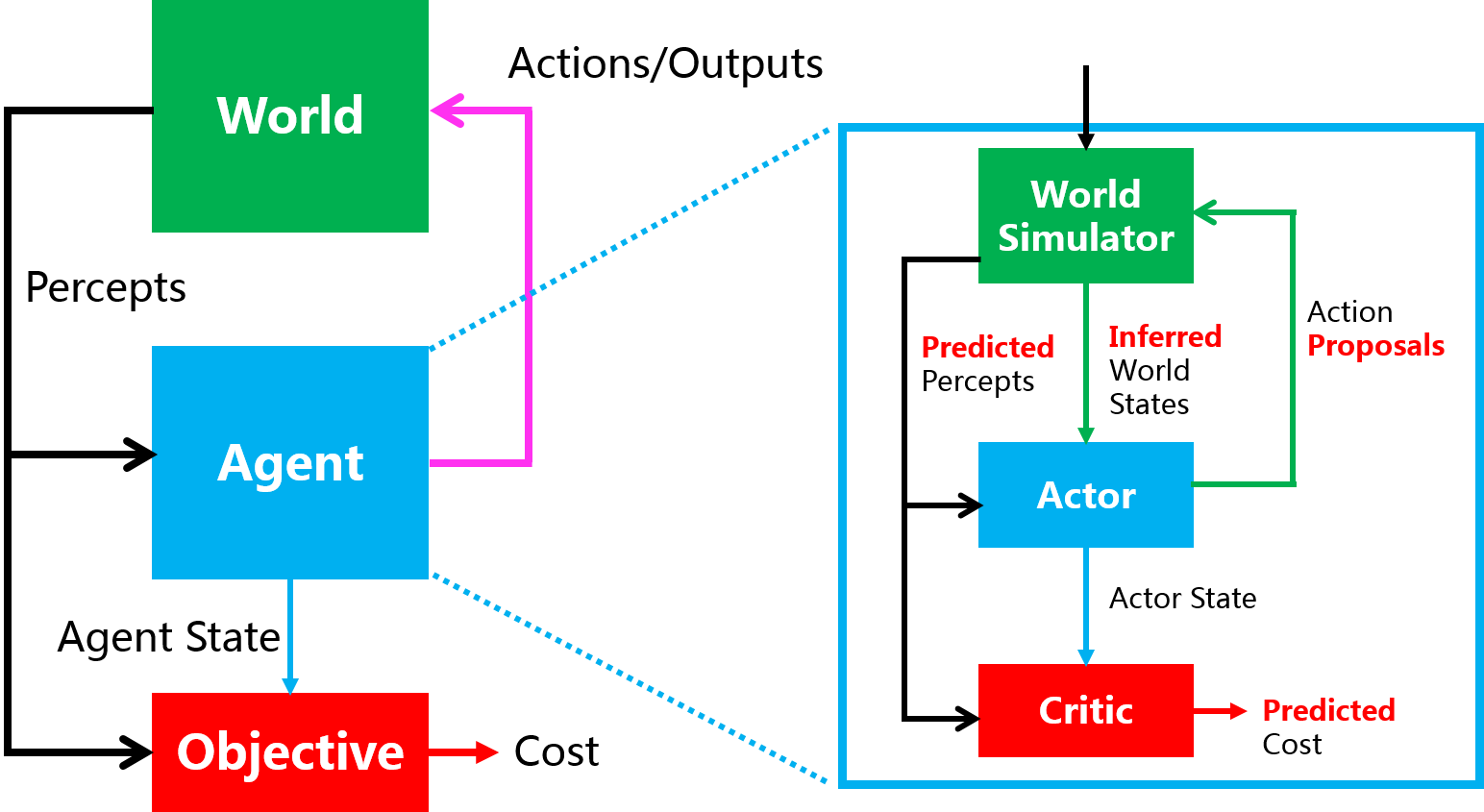}
\centering
\caption{LeCun's Agent Model with an Internal World Model[6]}
\label{fig4}
\end{figure}

\section{UNIFYING THE LEARNING AGENT MODELS}
The models proposed by Holland[2][3], Minsky[4] and Lecun[5] may approach the design of a learning agent model for strong AI from different perspective and different emphases. With some non-difficult manipulation of their models, the performance element, learning element and critic design for an unified intelligent agent model may be proposed as follows:
\begin{itemize}
\item Performance Element Design: An aggregation of micro-agents described by simple reactive rules.
\item Learning Element and Critic Design: A collection of action modules for building, manipulating and evaluating higher-level (likely to be non-linear) representation of micro-agents’ behaviours.
\end{itemize}

However, it is not apparent to come up with a concrete and operable design for the Problem Generator Design yet. Here, we try to solve the problem based on the following basic understandings:
\begin{itemize}
\item Non-Complete Representation as Motives: Even though an agent have non-linear basis/generators for representing very complex interactions in the world, we may still face up to non-complete representations, with limited resources. Therefore, the agent may adopt a sub-optimal (or an action that cannot be proved optimal with limited resource) in the world.
\item Multiple-Trial in the World Model: Though the agent cannot find the optimal actions, the agent may come up with a set of sub-optimal candidate actions, where the size of the set is limited based on available resources. After taking previous actions in the world, these candidate actions could be tested within the updated world model.
\end{itemize}

Based on the above insights, we may propose an operable pathway to the problem generators for strong AI.
\begin{itemize}
\item Problem Generator Design: A set of modules to evaluate non-complete representation of complex action-reward interactions, to generate candidate action set, and to trigger reflective actions in the world model..
\end{itemize}

We may now graphically represent our UIC model with the receivers for modelling sensors’ perceptions, transmitters for modelling actuators’ actions, as well as the performance element, learning element, critic and problem generator for information source/destinations. Please note we not only merge the different components in the learning agent model, but also provide guidelines for designing each of them.

The details of the UIC design would be provided in the following sections, through more elaborations, case studies and insightful discussions.

\begin{figure}[ht]
\includegraphics[width=0.35\linewidth]{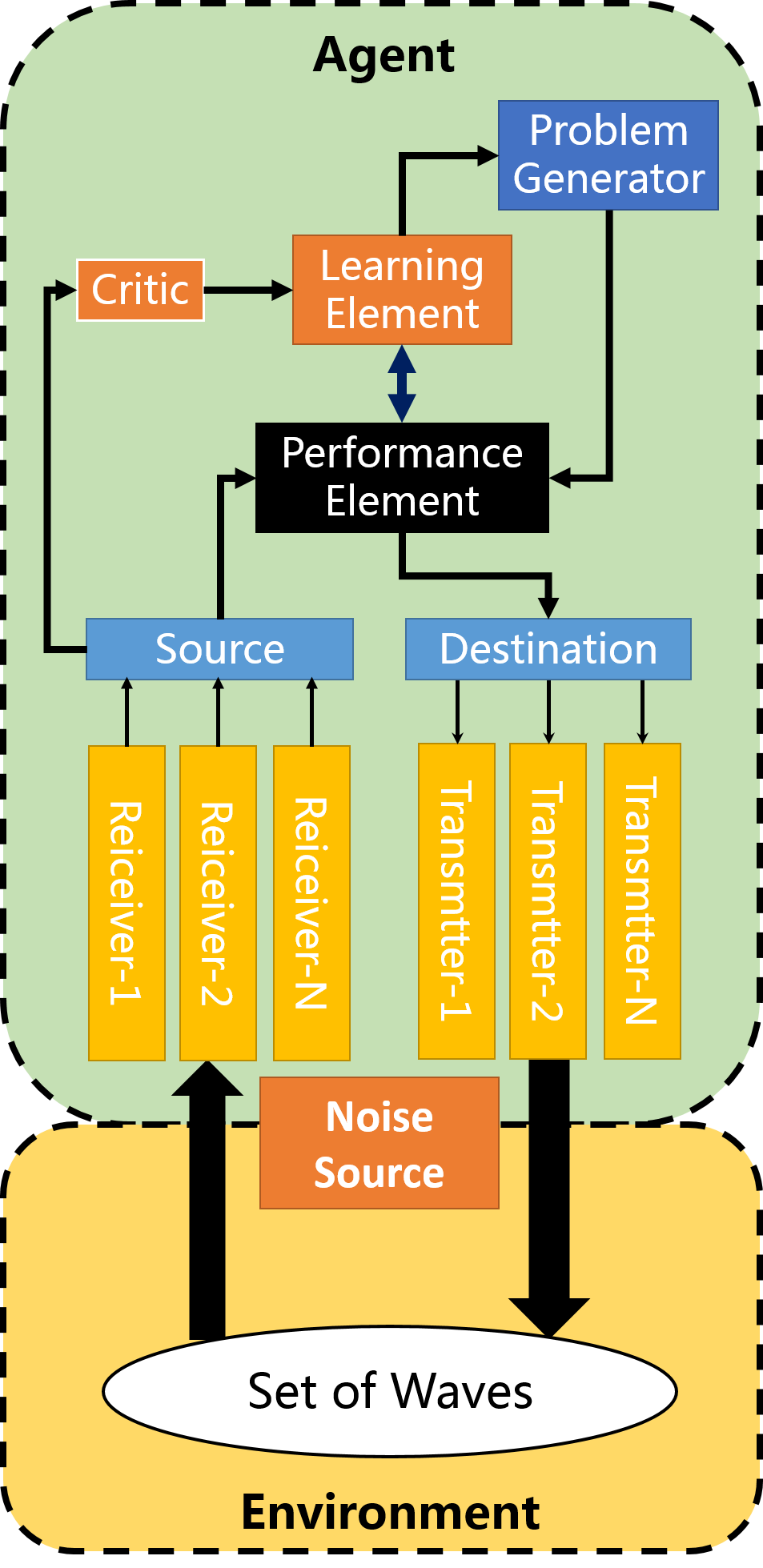}
\centering
\caption{The Proposed UIC model with an Internal Model for Learning Agent}
\label{fig5}
\end{figure}

\section{CONLUSIONS}
In this paper, we first review the Shannon model for communications, and propose an unified intelligence-communication model to describe the interactions between intelligent agents and the environment, where the perception and action of an agent may be modelled with information reception and transmission.

Then, we review and discuss the relationship between different models for intelligent agents with an emphasis on the learning agent model, and propose an unified, complete and operable design for intelligent agents with learning capabilities. Along with the perception and action model proposed, we may form the complete UIC model for an intelligent agent.

In the second part of this treatize, we would show that case studies would be provided, exemplifying the adoption of the UIC model. Specifically, we would dive deep into two representative cases, the analysis of a natural bee-colony, as well as the co-design of communication, perception and action in artificial multi-robot system.

\end{document}